\def\year{2021}\relax
\documentclass[letterpaper]{article} 
\usepackage{aaai21}  
\usepackage{times}  
\usepackage{helvet} 
\usepackage{courier}  
\usepackage[hyphens]{url}  
\usepackage{graphicx} 
\def\UrlFont{\rm}  
\usepackage{natbib}  
\usepackage{caption} 
\usepackage{amsfonts}
\usepackage{amsmath}
\usepackage{caption}
\usepackage{subcaption}
\usepackage{booktabs}
\usepackage{listings}
\usepackage{xcolor}
\usepackage{xspace}
\usepackage{multirow}
\usepackage{makecell}
\usepackage{algorithm}
\usepackage{algpseudocode}

\definecolor{codegreen}{rgb}{0,0.6,0}
\definecolor{codegray}{rgb}{0.5,0.5,0.5}
\definecolor{codepurple}{rgb}{0.58,0,0.82}
\definecolor{backcolour}{rgb}{0.95,0.95,0.92}
\usepackage{amsthm}

\lstdefinestyle{mystyle}{
    backgroundcolor=\color{backcolour},   
    commentstyle=\color{codegreen},
    keywordstyle=\color{magenta},
    numberstyle=\tiny\color{codegray},
    stringstyle=\color{codepurple},
    basicstyle=\ttfamily\footnotesize,
    breakatwhitespace=false,         
    breaklines=true,                 
    captionpos=b,                    
    keepspaces=true,                 
    numbers=left,                    
    numbersep=5pt,                  
    showspaces=false,                
    showstringspaces=false,
    showtabs=false,                  
    tabsize=2
}

\newcommand{\traindata}{\mathcal{D}\xspace} 
\newcommand{\lissa}{\texttt{LISSA}\xspace}
\newcommand{\tracin}{\texttt{TracIn}\xspace}
\newcommand{\randproj}{\texttt{RandProj}\xspace}
\newcommand{\randselect}{\texttt{RandSelect}\xspace}
\newcommand{\arnoldi}{\texttt{Arnoldi}\xspace}

\newcommand{\githuburl}{\url{https://github.com/google-research/jax-influence}}

\title{Scaling Up Influence Functions}

\author {
    Andrea Schioppa\thanks{Google AI Resident.},
    Polina Zablotskaia,
    David Vilar, 
    Artem Sokolov\\
}
\affiliations {
    Google Research\\
    \{arischioppa, polinaz, vilar, artemsok\}@google.com
}

\newtheorem{lemma}{Lemma}

\begin{document}

\maketitle

\begin{abstract}
We address efficient calculation of influence functions~\cite{koh2017understanding} for tracking predictions back to the training data. We propose and analyze a new approach to speeding up the inverse Hessian calculation based on Arnoldi iteration~\cite{arnoldi}. With this improvement, we achieve, to the best of our knowledge, the first successful implementation of influence functions that scales to full-size (language and vision) Transformer models with several hundreds of millions of parameters. We evaluate our approach on image classification and sequence-to-sequence tasks with tens to a hundred of millions of training examples. Our code will be available at \githuburl.
\end{abstract}

\noindent 

\section{Introduction}

Recognizing data's highest agency in defining deep neural networks' (DNNs) performance, the pursuit of state-of-the-art has made datasets for training modern DNNs grow to sizes that can no longer be curated by humans. This has acutely aggravated data issues like noise and mislabeled data: Noise is characteristic of tasks where training data is crawled from the Web (e.g.~machine translation) and where golden-truth labels are heuristically paired to inputs~\cite{uszkoreit2010large}, leaving ample room for errors and inheriting biases of the heuristic. Wrong labels can be also introduced 
by non-expert crowd annotators who, considering the amount of data to be labeled, are hard to incentivize for quality within available budgets~\cite{bowman}. 


Given the above, a natural way to interpret and fix DNN models is to track their bad (or good) predictions down to the training examples that caused them~\cite{cook80,koh2017understanding, yeh}, and take appropriate action on the found examples or annotation policies. 
%
Addressing this, \citet{koh2017understanding} proposed \emph{influence functions} (IFs) as a theoretically motivated method, grounded in robust statistics~\cite{cook88}, of quantifying the effect of training examples on predictions: For a query example $z$, IFs estimate the most influential example $x$ in training data $\traindata$, in terms of absolute change of loss $L$ if $x$ were infinitesimally up-weighted in $\traindata$, with: 
\begin{equation}\label{eq:hessian_influence}
\mathcal{I}_H(x, z) = \langle \nabla_{\Theta}L(z), H^{-1} \nabla_{\Theta}L(x)\rangle,
\end{equation}
where $H=\nabla^2_\Theta L$ is the Hessian of the model at parameters $\Theta$.
The straight-forward IF implementation, using the approximate Hessian inversion procedure \lissa~\cite{lissa_paper}, has $O(p)$ memory and 
$O(r\cdot p)$ 
time complexities, where $r$ is the \lissa iteration count 
and $p=|\Theta|$, incurred at every $x$. Besides the need of careful tuning of \lissa, the $O(p)$-memory has been the major obstacle on the way of IF deployment for debugging application-relevant DNNs with \emph{hundreds of millions} (or more) training examples and model parameters; so
noise or mislabeling issues remain unfixed or even undetected, and are adversely impacting predictions.

In this work, we focus on reducing the IF memory footprint by not materializing $O(p)$-size gradients nor Hessians, and decoupling the required number of $H$ estimations, \mbox{$O(r\cdot |\traindata|)$}, from the training data size. This allows to parallelize computation over larger $b$ and scale to huge datasets and models. Specifically, we use Arnoldi iteration~\cite{arnoldi} to find the dominant (in absolute value) eigenvalues of $H$ and their orthonormal eigenvectors on a random data subset, $|\traindata'|\ll|\traindata|$, and then cheaply invert the diagonalized $H$, avoiding calls to \lissa as well as its convergence and stability issues. As $H$ is Hermitian, ~\eqref{eq:hessian_influence} is symmetric w.r.t.~$x$ and $z$, so previous work cached $\{\nabla_\Theta L(x)\}$ to improve IFs usability~\cite{fast_if}, however, it only spares one backward pass per $x$ and requires to re-estimate the product of the (unstorable) $H^{-1}$ with $\nabla_\Theta L(z)$ every time an influence on $z$ is requested. The crux of our approach is in caching instead $H$ in the trivially-invertable diagonalized form and for the small-dimensional subspace spanned by a few dominant eigenvectors, $\tilde p \ll p$. 
Hessian-gradient products are then reduced to simple scalar-gradient products, which do not need to be materialized in memory as~\eqref{eq:hessian_influence} can now be implemented with Jacobian-vector products. In summary, our approach renders repeated re-estimations of $H^{-1}\nabla_\Theta L(x)$ at every $x$ unnecessary and they are replaced with the memory- and time-efficient forward-mode differentiation. 

Empirically, IFs with Arnoldi iteration achieve speed-ups of 3-4 orders of magnitude over the \lissa-powered IFs~\cite{koh2017understanding} and of 10x over \tracin~\cite{tracin}, a heuristic gradient-only alternative to IFs (\S\ref{sec:experiments}), with better or similar accuracy. With this improvement, we successfully evaluated IFs on both language and vision, full-size Transformer models (up to 300M of parameters) in image classification and sequence-to-sequence tasks, resp., on 14M (\mbox{ImageNet}) and 100M (Paracrawl) training examples. 

Note that the standard conditions for~\eqref{eq:hessian_influence} to be a correct influence estimate, i.e.~locally strictly-convex $L \in \mathcal{C}^2$~\cite{koh2017understanding}, remain in place and their fulfilment depends on the concrete task, network, training algorithm and its convergence status. The time/memory complexity also remain, however, our contribution improves constants hidden in the $O$-notation, and thus permits IF evaluation on full data/models that are relevant in applications, on standard memory-limited hardware. This opens the way for developers to make informed decisions if IFs are appropriate for their task, rather than to resort to heuristics from the start. This is encouraging, since the existing brute-force recipe of ``soothing'' the $O(p)$ complexity by subsetting parameters, e.g.~focusing on few layers only~\cite{multistage}, is prone to producing incorrect results~\cite{feldman_practical} (see also \S\ref{sec:mt}). On the other hand, running IF on subsets of $\traindata$ to reduce runtime~\cite{fast_if} may introduce unwanted biases, misattribute prediction failures, and would not be enough for identifying mislabeled examples~\cite{tracin} or ``hard'' examples requiring memorization~\cite{feldman_practical}. Yet, these approaches are compatible with our method and should result in compound speed-ups.

We will open-source our implementation of Arnoldi iteration at \githuburl.

\section{Related work}

Explaining DNN predictions falls under a broader interpretability umbrella, where the lingering complexity of the  data-explainability approach made research historically focus on instance-based methods, that explain predictions in terms of task-specific structural units of inputs, e.g.~pixels or tokens. Rich literature offers different instantiations of the idea: gradient-based saliency maps~\cite{simonyan2013deep},
input perturbations~\cite{LiMJ16a} or LIME~\cite{ribeiro2016should}, which fits a linear model in the inputs neighborhood. However, being limited to specific inputs, their insights are rarely actionable for system developers. And while it is possible to repurpose them to data explainability, e.g.~via clustering of saliency maps~\cite{muller}, this solves a more difficult task than necessary, introduces new hyperparameters (incl.~the saliency method itself) and relies on human experts to make sense of the clusters. 


In contrast to instance-explanations in the form of token level heatmaps, the IF provides a method for tracing model predictions back to training examples. Existing approaches to reducing IF runtime mostly address salient problem axes -- dimensionality of active parameters, cardinality of data subset or number of iterations -- without addressing the procedure itself; or they drop theoretical foundations to use heuristics simpler than IF: e.g.~\citet{tracin} reduce influence to tracking loss changes with the cumulative (over training model checkpoints, i.e.~model snapshots) dot products of gradients, and in~\cite{yeh} authors leverage kernel functions evaluated at the training samples for explaining inference decisions. 


Mathematically, the closest to our work is~\cite{spectrum_guys} who use a specialization of Arnoldi iteration to Hermitian matrices (Lanczos iteration) to study the dynamics of spectra of entire Hessians at different snapshots during training. Because of this different goal, they use full-batch Hessians (i.e.~computed on the full dataset), while we ``spread'' the Hessian approximation across smaller batches that do not cover the full $\traindata$. In result, we can work with larger models and datasets, e.g.~ResNet50/ViT vs.~ResNet18, and at a larger speed (simultaneously bumping the number of Lanczos/Arnoldi iterations from 90 to 200 to increase precision).

\section{Influence and Influence Functions}

The \emph{true} influence of $x$ on $z$, $\mathcal{I}_{\rm true}(x, z)$, is defined as the change in the loss $L$ at $z$
between having learned the model parameters $\Theta$ without and with $x$ in the training data~\cite{cook80}:
$$
\mathcal{I}_{\rm true}(x, z) = L(z | \Theta: x\not\in \traindata) - L(z | \Theta: x\in \traindata).
$$
Explicit calculation of $\mathcal{I}_{\rm true}(x, z)$, by removing every $x$ and retraining, is infeasible for large $\traindata$ and several approximation techniques have been proposed. For example,
\citet{feldman_theory, feldman_practical} propose to train multiple models on randomly selected data subsets while tracking, for each $x$,
to which subsets it belonged; this way, one can obtain an unbiased estimator $\mathcal{I}_{\rm mem}(x,x)$ of $\mathcal{I}_{\rm true}(x, x)$ which, however, requires a substantial amount of model re-trainings (up to thousands to satisfy theoretical guarantees~\cite{feldman_theory}). 

\citet{koh2017understanding} advocated the use of IFs in~\eqref{eq:hessian_influence} that approximate the loss change after an infinitesimal up-weighting of~$x$ in $\traindata$.
For models used in practice, $H$ cannot be materialized in memory, let alone be inverted by standard linear algebra. However, for a fixed 
vector $v$, the Hessian-vector product (HVP), $H v$, can be computed in $O(b\cdot p)$ time and 
memory~\cite{pearlmutter}, where 
$b$ is the batch size and defines the number of training examples on which $H$ (of an implicitly given loss $L$) will be approximated. HVP is commonly implemented in modern autodiff toolkits~\cite{baydin2015automatic} as the reverse-mode differentiation Jacobian-vector product (JVP), followed by a forward-mode JVP. 

Repeated HVP calls are the workhorse of the iterative procedure \lissa~\cite{lissa_paper}, used by~\cite{koh2017understanding}, that estimate inverse HVP as:
$$
H^{-1}_r v = v+(I-H)H^{-1}_{r-1} v,
$$
where $H$ is approximated on random batches and $v$ is a gradient. Even for small $r$, the procedure is both time- and memory-expensive as the $O(p)$-memory of HVP on explicitly instantiated $v$ forces to estimate $H$ on a \emph{single sampled training point per iteration} ($b=1$), impacting accuracy. Moreover, the total $O(r\cdot b\cdot p)$ time complexity will be incurred at \emph{every} $x$ in whose influence on $z$ we are interested. 


\paragraph{Evaluating influence methods.}\label{sec:eval}
A practical problem with influence-based explainability is the absence of ground-truth to verify that a method produces correct results. In this paper we use two proxies, following the assumption that $x$s with high self-influence $\mathcal{I}_H(x,x)$ correspond to data outliers~\cite{koh2017understanding, tracin}: we either introduce a known synthetic data corruption and check for its correct retrieval by a method, or filter high-influence points out and measure a change in downstream task metrics~\cite{fast_if,kocijan20}.
Using $\mathcal{I}_H(x,x)$ as a retrieval score for corrupted data, we measure the retrieval quality as areas under the ROC and the precision-recall curves, respectively denoted as the Area Under the Curve (AUC) and the Average Precision (AP). 

\section{Scaling Influence Functions}\label{sec:scaling}

From the discussion above, the $O(p)$-memory complexity is the major bottleneck for efficient implementation of IFs. 
We start with an overview of existing techniques, showing their commonalities. Note that all are compatible with the approach we propose.

\paragraph{Caching.} For data-interpretability purposes one might consider limiting~\eqref{eq:hessian_influence} to a sufficiently promising subset $\traindata'\subset \traindata$, e.g. \citet{fast_if} define $\traindata' $ as the top\nobreakdash-$k$ $\ell_2$-neighbors of $z$ in $\traindata$. Besides more hyperparameters, this still requires computing $H^{-1}\nabla_\Theta L(z)$ for every $z$ and, as discussed above, reducing $\traindata'$ would not be enough for applications that require computing $\mathcal{I}_H(x,z)$ on \emph{all} training $x$.
As $H$ is symmetric, one could swap $x$ and $z$, and cache $H^{-1}\nabla_\Theta L(x)$ instead, bringing down the query complexity 
having only
to compute $\nabla_\Theta L(z)$ now, but this would just shift the computational burden to building the search index over $H^{-1}\nabla_{\Theta}L(x)$.

\paragraph{Restricting parameters.} Reducing required memory is possible by naively limiting the computation to a smaller subset of parameters of cardinality $\tilde{p}$, e.g.~selecting one or few layer(s); usually, the last layer is selected~\cite{koh2017understanding}. This has two drawbacks: the choice of layers becomes a hyperparameter and the viable values of $\tilde{p}$ will depend on the model architecture, and as~\citet[\S 3.6]{feldman_practical} show using just one layer can result in different influence estimates compared to the full model.

\paragraph{Random projections.}
For simplification one might assume $H=I$ and reduce influence estimates to dot products of gradients.
To account for multiple layers at once and to get a finer-step control of $\tilde p$, we consider a simple baseline, \randselect, which randomly
selects $\tilde{p}$ parameters $\tilde{\Theta}\subset\Theta$ and computes influence using the final checkpoint
and gradients with respect to $\tilde{\Theta}$, and can be combined with layer selection. The \randselect estimator
can be equivalently expressed as
\begin{equation}\label{eq:projected_estimate}
\mathcal{I}_G(x,z) = \langle G\nabla_{\Theta}L(x), G\nabla_{\Theta}L(z)\rangle,
\end{equation}
where $G\in \mathbb{R}^{\tilde{p}\times p}$ is a row selection matrix of the gradient's components corresponding to~$\tilde{\Theta}$.

We also use another sketching~\cite{sketching} baseline, \randproj, initially proposed by~\citet{wojnowicz} for generalized linear models: for a random Gaussian projection matrix $G$, $\mathbb{E}[G^TG]=I$, which leads to an unbiased estimate in~\eqref{eq:projected_estimate}. Since normally $\tilde p \ll p$, it allows a memory-efficient implementation in the forward mode: one just estimates $\tilde p$ JVPs with the rows of $G$ that avoid materializing $O(p)$-size gradients. This has
a lower memory footprint than \randselect, which requires back-propagation as its $\tilde p$ (e.g.\ one layer) is still of the same order as $p$ (see \S\ref{sec:mt}).

\paragraph{Tracing updates.}
A development of the $H=I$ idea is proposed in \cite{tracin}, where influence approximation works for the case when one can trace changes to the loss $L$ across \emph{all} gradient steps. In order to make this feasible, they propose the \tracin estimator, defined on a subset of gradient steps:
$$
\mathcal{I}_{\rm TracIn}(x, z) =  \frac{1}{C}\sum_{i=1}^C\langle\nabla_{\Theta_i}L(x), \nabla_{\Theta_i}L(z)\rangle,
$$
where the $\{\Theta_i\}$ is a set of $C$ checkpoints. Note that the complexity of \tracin is $C$ times that 
of using exact gradient similarity and, as discussed in~\cite{tracin}, care needs to be taken in selecting checkpoints.
Another practical obstacle to \tracin is that, when analysing publicly released models, usually only the final model checkpoint is provided.

\paragraph{Compatible projections.}
\randproj assumes the full-dimension Hessian, $H=I$; if we drop this requirement, we might consider $H$ restricted to the 
subspace $S_G$ which is the image of $G$, and work with $G\cdot H\cdot G^T$ instead of the larger $H$.
However, $S_G$ is not in general $H$-invariant which can lead to approximation errors as when $H$ is applied
to a vector $v\in S_G$, the result might have non-negligible components orthogonal to $S_G$. We will see an example
of this later in the experiments on eigenvalue retrieval for MNIST (Figure~\ref{fig:mnist_eigenvalue_retrieval})
where \randproj requires a considerably larger $\tilde p$ than \arnoldi to retrieve the top-$\tilde p$
eigenvalues of $H$.

\paragraph{Our approach.} We propose to use the standard technique of building an approximately $H$-invariant subspace by selecting an arbitrary (e.g.~random) vector $v\in\mathbb{R}^p$ and constructing the $n$-th order Krylov subspace:
$
K_{n}(H; v) = {\rm Span}\{ v, Hv, H^2 v, \dots, H^n v\}.
$
The Arnoldi iteration~\cite{arnoldi} additionally builds an orthonormal basis for $K_n(H; v)$, so that the diagonalization of the restriction $\tilde H$ of $H$ to $K_{n}(H; v)$ yields an approximation of the largest (in absolute value) eigenvalues of $H$ and of the corresponding eigenvectors~\cite[Ch.~33-34]{trefethen97}. Assuming $n$ is large enough to estimate the largest $\tilde{p}$ eigenvalues, in summary we obtain a projection matrix $G$ and work with $\tilde H = G\cdot H \cdot G^T$, which is a smaller dimensional matrix. We will call this algorithm \arnoldi, with the pseudocode in Algorithm~\ref{alg:arnoldi}. 

The common feature of \randproj and \arnoldi is that, instead of working with the full gradient $\nabla_\Theta L(x)$, one takes the JVPs $\langle g_i, \nabla_\Theta L(x)\rangle$ with respect to the rows $g_i$ of~$G$.
The implementation then becomes considerably more efficient as it can be done in the forward-mode differentiation and 
on larger batches. Moreover, in the case of \arnoldi the matrix $H$ gets replaced with now diagonal $\tilde H$, simplifying the matrix inversion appearing in the definition of $\mathcal{I}_H$, and dispensing with the expensive \lissa procedure.

\paragraph{Error analysis.} It remains to analyse the effect of using top-$\tilde p$ eigenvalues in \arnoldi. Recall that~\citet{koh2017understanding} derive~\eqref{eq:hessian_influence} by minimizing the quadratic form 
$
Q(\theta) = \frac{1}{2}\langle\theta, H\theta\rangle -\frac{1}{N}\langle\nabla_\Theta L(x|\Theta_0),\theta\rangle,
$
where $\Theta_0$ are the parameters at convergence. Ordering the eigenvalues of $H$ at $\Theta_0$, $|\lambda_1| \ge |\lambda_2| \ge \cdots$
and letting $e_1, e_2, \cdots$ be the corresponding eigenvectors, \citet{spectrum_guys} empirically observe (and prove in the quadratic case) that gradient updates align with the subspace of $H$ corresponding to the dominant $\lambda$s. 
We provide two additional arguments in the same direction: we upperbound the error of approximating
$Q$ using such a subspace in Lemma~\ref{lmm:error}, and discuss the effect of noise in $H$ and
the size of $\lambda_k$ on applying $H^{-1}$ to a vector in Lemma~\ref{lmm:var} (with proofs in~\S\ref{app:approximation}). 

Let $Q_k$ be the form $Q$ restricted to the $H$-subspace spanned by the top-$k$ $\lambda$s. We show that, as $k$ increases, $Q_k$ approximates $Q$ better and the errors in directions of $e_k$ corresponding to smaller $|\lambda_k|$ matter less\footnote{One might attempt 
\arnoldi on $H_{\Theta_0}^{-1}$ to obtain an approximation directly in the
subspace of the top-$k$ eigenvalues of $H_{\Theta_0}^{-1}$. We found this approach however to be less performant (see \S\ref{app:inverse_arnoldi}).}:

\begin{lemma}\label{lmm:error}
$Q_k$ approximates $Q$ by an error bounded by:
$
0\le Q(\theta) - Q_k(\theta) \le  \frac{1}{2}|\lambda_{k+1}| \|\theta\|_2^2.
$
Further, if minimizing $Q$ introduces an error $\varepsilon$ in the direction of $e_{k+1}$ obtaining an estimate $\theta'$ for $\theta_*$, then 
$
Q(\theta')-Q(\theta_*)=\frac{\varepsilon^2}{2}\lambda_{k+1}.
$
\end{lemma}

Another way of looking at the same phenomenon is to consider the variance of estimated influence as the function of $|\lambda_k|$. Consider a computation of $y = H^{-1}u$, where vector $u$ is known exactly. Assume also that $H$'s estimation is noisy resulting in error $H+\delta H$, that  $\mathbb{E}[\delta H]=0$, and that the $\delta H$ is isotropic (e.g.~does not preferentially align with some $e_k$ nor co-vary with $\lambda_k$). 
Then the variance of the estimator $\hat y = (H_{\Theta_0}+\delta H)^{-1}u$ in the direction of $e_k$ is proportional to~$|\lambda_k|^{-2}$:
\begin{lemma}\label{lmm:var}
The variance of $\hat y$ in the direction of $e_k$ is 
$
{\rm Var}(\langle\hat y, e_k\rangle) \approx \frac{1}{|\lambda_k|^2}{\rm Var}(\langle \delta H e_k, y\rangle).
$
\end{lemma}


\begin{algorithm}[t]
\algtext*{EndFor}
\algtext*{EndProcedure}
\caption{\arnoldi}\label{alg:arnoldi}
\begin{algorithmic}[1]
\small
\Procedure{Arnoldi}{$v$, $n$}
\\\Comment{Build orthonormal basis for the Krylov subspaces $K_n$.\mbox{~~~~~~~~~}}
\State $w_0\gets \frac{v}{\|v\|_2}$
\State $A_{l,m}\gets0$ for $0\le l\le n$ and $0\le m < n$
\For{$i\gets 1,n$}
\State $w_i\gets H \cdot w_{i-1} $ \Comment{HVP in fwd-over-rev mode}\label{line:hvp}
\State Set $A_{i,j}=\langle w_i, w_j\rangle$ for $j<i$
\State  $w_i\gets w_i-\sum_{j<i}A_{i,j}w_j$ \Comment{Orthogonalize}
\State $A_{i+1,i}\gets\|w_i\|_2$, $w_i\gets\frac{w_i}{\|w_i\|_2}$
\EndFor
\State \textbf{return}: {$A, \{w_i\}$}
\EndProcedure
\Procedure{Distill}{$A, \{w_i\}, \tilde p$}
\\\Comment{Distill $A, \{w_i\}$ to its top-$\tilde p$ eigenvalues.\mbox{~~~~~~~~~~~}\mbox{~~~~~~}\mbox{~~~~~~~~~~~~~~~}}
\State Discard the last row of $A$ and the last $w_n$
\State Obtain $A$'s eigenvalues $\{\lambda_i\}$ and eigenvectors $\{e_i\}$
\State Set $\{\lambda'_i\}$ to the $\tilde p$-largest (in absolute value) of $\{\lambda_i\}$ 
\State Set $\{e'_i\}$ to the corresponding eigenvectors
\State Set $G$ to the projection onto the spans $\{e'_i\}$ in  $\{w_i\}$-basis 
\State \textbf{return}: $\{\lambda'_i\}$, $G$
\EndProcedure
\Procedure{Influence}{$x, z, n, \tilde p$}
\\\Comment{Influence of $x$ on $z$ with $n$ iterations and top-$\tilde p$ eigenvalues.}
\State $v\gets \mathcal{N}(0,1)$ 
\State $A$, $\{w_i\}$ = \textsc{Arnoldi}($v, n$)%
\rlap{\smash{~~~~~~~~~$\left.\begin{array}{@{}c@{}}\\{}\\{}\end{array}\right\}$\Comment{\begin{tabular}{cc}Executed once\\and cached\end{tabular}}}}
\State $\{\lambda'_i\}$, $G$ = \textsc{Distill}($A$, $\{w_i\}$, $\tilde p$)
\State $g_x\gets G\cdot \nabla_\Theta L(x)$\Comment{fwd JVP for $x$ over $G$-rows}
\State $g_z\gets G\cdot\nabla_\Theta L(z)$\Comment{fwd JVP for $z$ over $G$-rows}
\State $g_x\gets g_x/\{\lambda'_i\}$ \Comment{Multiply with diagonalized $\tilde H^{-1}$}
\State \textbf{return}: $\langle g_x, g_z\rangle$
\EndProcedure
\end{algorithmic}
\end{algorithm}


\section{Experiments}\label{sec:experiments}
\subsection{Small Model \& Data Scale: Digit Recognition}\label{sec:mnist}
In this subsection, to be able to compare all baselines we pick the small MNIST dataset~\cite{mnist} and consider two CNNs of different sizes: a small one that permits the exact Hessian calculation, and a larger one on which we can gauge the scalability potential.

Because the influence calculation with \lissa and \tracin is slow, following~\cite{koh2017understanding}, we take two $10\%$ subsamples of the original data for training and evaluation, and randomly relabel $20\%$ of training examples to create a corrupted dataset to evaluate mislabeled example retrieval with influence estimates. Unlike~\cite{koh2017understanding, tracin} we introduce the noise \emph{before} training the models; by design, a perfect model on correctly labeled data would achieve only $80\%$ accuracy on our eval set.

\paragraph{Small network.}
We re-implemented the small convolutional network with smooth non-linearities from~\cite{koh2017understanding}, and trained it following their recipe that is designed to make the assumptions behind the influence function method~\cite{cook88} satisfied: First, to ensure convergence, the network is trained for more steps (500k) than one would normally do, with a large batch size of 500 images. Second, the $\ell_2$-regularization of $5\cdot 10^{-3}$ is introduced to make $H$ positive definite. With only 3k parameters, it is a useful benchmark where it possible to compute $H$ explicitly. We achieve accuracy of 73.8\% and 92.3\% on, resp., the corrupted and the true evaluation set.

\begin{figure}
\centering
    \includegraphics[width=0.9\columnwidth]{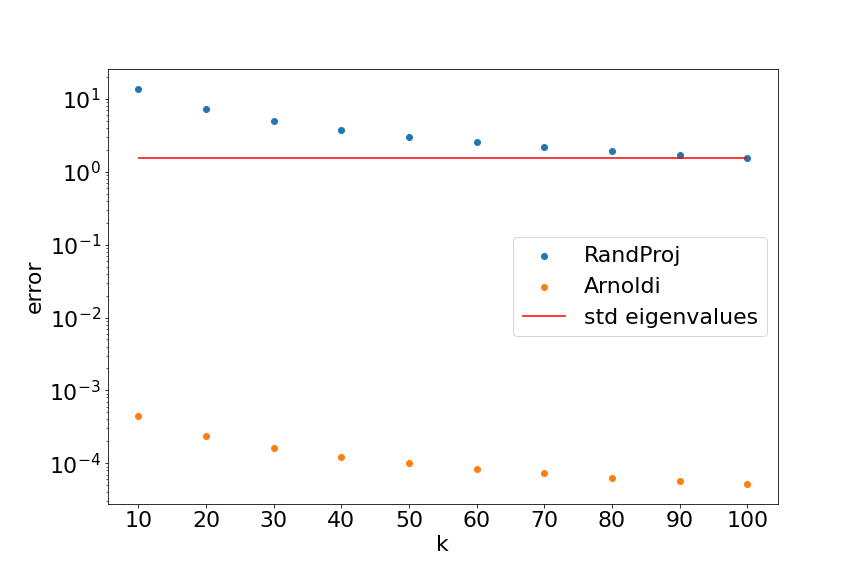}
\caption{Estimation error of the top-$\tilde p$ eigenvalues obtained by \textsc{Distill} as the optimal transport (Wasserstein) distance to the exact $H$ eigenvalues (assuming uniform distribution). 
The horizontal line is the standard deviation of the latter.}

\label{fig:mnist_eigenvalue_retrieval}
\end{figure}

In Figure~\ref{fig:mnist_eigenvalue_retrieval} we compare the accuracy between the top\nobreakdash-$\tilde p$ eigenvalues 
estimations obtained by \arnoldi (for $n=200$) and \randproj. This illustrates the point made above 
that if the image subspace $S_G$ associated with the 
projection $G$ is not $H$-invariant, then eigenvalue estimates can be poor.

\begin{figure*}[tb!]
\begin{subfigure}{\columnwidth}
\centering
\includegraphics[width=0.9\columnwidth]{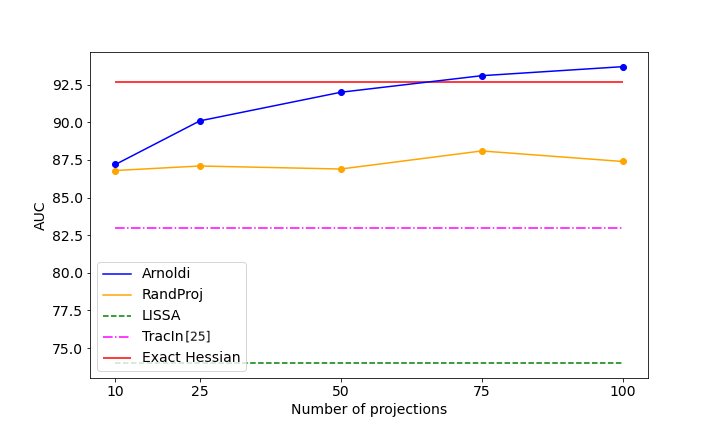}
\end{subfigure}
\begin{subfigure}{\columnwidth}
\centering
\includegraphics[width=0.9\columnwidth]{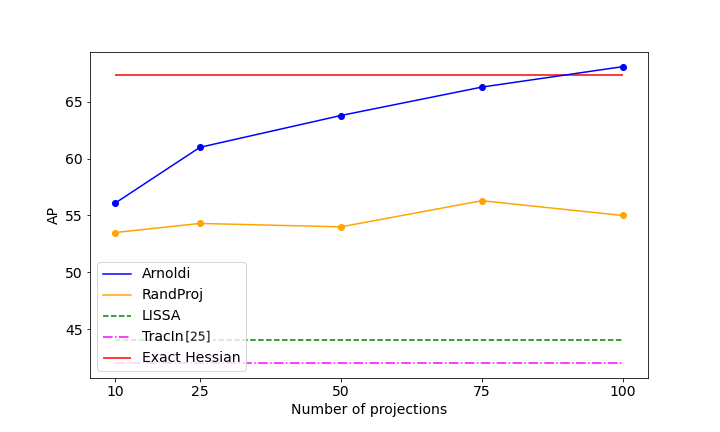}
\end{subfigure}
\caption{AUC and AP for retrieval of mislabeled MNIST examples as a function of $\tilde p$ for the small CNN model.}
\label{fig:mislabel-mnist-retrieval}
\end{figure*}
In Figure~\ref{fig:mislabel-mnist-retrieval} we plot the retrieval quality of mislabeled examples
by \arnoldi and \randproj as a function of $\tilde p$. The horizontal lines correspond to using the exact Hessian, \lissa and \tracin ($C=25$ checkpoints, taken every 10 epochs). For this network we see that \arnoldi outperforms \randproj and steadily improves for further larger values, outperforming even the exact Hessian for a large enough $\tilde p$ (which can be explained by presence of close-to-zero eigenvalues in the exact $H$ which affects its inverse).

\paragraph{Larger network.}
To consider a more realistic and larger network we use the CNN from the Flax library\footnote{https://github.com/google/flax/tree/master/examples/mnist}
with 800k parameters, still small by industry standards, but for which $H$ cannot be already computed explicitly. We train it for 10 epochs on GPU V100 without regularization, and achieve 75.4\% on the corrupted and 94.8\% on the true labels test set. This network is more realistic in size and in training procedure than the one in~\cite{koh2017understanding}.

Table~\ref{tab:mislabel-mnist-retrieval} reports results of retrieval of mislabeled examples with the total scoring time $T$ as there is a trade-off between gains in AP or AUC vs.~$T$. As computing exact $H$ was not possible, our baseline for self-influence scoring is \lissa, which is about $10^4$ times slower than \arnoldi (which took 353 sec to estimate $H$ for $\tilde p=10$, $n=200$ and $b=512$).

We find that both \arnoldi and \randproj have a good trade-off between retrieval accuracy and speed, while
\randselect, despite being the fastest, suffers in retrieval quality. As here \randproj
performs slightly better than \arnoldi, we initially hypothesized that this might indicate an insufficient accuracy of HVP estimations: To verify, we re-ran \arnoldi with the HVPs estimated on the \emph{full} dataset and obtained almost identical results, indicating that, on the one hand, \arnoldi accurately estimates eigenvalues with only 512 examples per HVP and, on the other hand, for this network and task, gradient similarity methods might be more appropriate (cf. \lissa for $r=100$ does not move the needle either). Finally, \tracin on 10 checkpoints (after each epoch) had the best retrieval quality, however, while its runtime on this model and dataset is acceptable, it becomes a problem for the larger models we consider later.

In \S\ref{app:mnist_oppo}, we show that images with $\mathcal{I}(x,z)<0$ do appear ambiguous to humans or are incorrectly labeled. Results for the same CNN when over-trained or regularized are in~\S\ref{app:longer_cnn}.


\begin{table}[t]
    \centering
   \resizebox{0.9\columnwidth}{!}{
    \begin{tabular}{lrrrr}
\toprule
  Method &    $\tilde{p}$ &    $T$, secs & AUC & AP \\
\midrule
\lissa, $r=10$ & - & 4900 & 98.9 & 95.0 \\
\lissa, $r=100$ (10\% $\Theta$) & - & 32300 & 98.8 & 94.8 \\
\midrule
\tracin[1] & - & 5 & 98.7 & 94.0 \\
\tracin[10] & - & 42 & \textbf{99.7} & \textbf{98.7} \\
\midrule
\randproj & 10 & 0.2 & 97.2 & 87.7 \\
\randproj & 100 & 1.9 & 98.6 & 93.9 \\
\midrule
\randselect & 10 & 0.1 & 54.9 & 31.2 \\
\randselect & 100 & 1.8 & 91.8 & 72.6 \\
\midrule
\arnoldi & 10 & 0.2 & 95.0 & 84.0 \\
\arnoldi & 100 & 1.9 & 98.2 & 92.9 \\
\bottomrule
\end{tabular}}
    \caption{Retrieval of mislabeled MNIST examples using self-influence for larger CNN. For \tracin the
    $C$ value is in brackets (last or all). All methods use full models (except the \lissa run on 10\% of parameters $\Theta$). For \randproj std deviation of AUC/AP estimates is 0.1/0.7 over 20 runs.}
    \label{tab:mislabel-mnist-retrieval}
\end{table}

\subsection{Scaling with Data Size: Machine Translation}\label{sec:mt}
To test scalability over the data size dimension, we investigate IFs for machine translation focusing on data selection over millions of examples. We verify the hypothesis that the ``cleanest'' data is the one with the lowest self-influence~\cite{koh2017understanding, tracin}. We evaluate retrieval of artificially corrupted examples on the WMT17 dataset (6M sentences), and evaluate data cleaning on the large noisy Paracrawl corpus (100M sentences). 
\arnoldi used $b=512$ for HVPs; performing $n=200$ iterations took 139 minutes.

\paragraph{Retrieving mislabeled parallel data.}
We experiment with the Transformer Base model~\cite{vaswani2017all}, implemented in Flax\footnote{https://github.com/google/flax/tree/main/examples/wmt}, on the clean WMT17 dataset for the German-English direction. We follow the original setup from~\cite{vaswani2017all} and train it for 100k steps on a 16-core TPUv2 (details in \S\ref{app:mt_details}). As is standard in machine translation, the model is neither over-trained to full convergence nor we employ the $\ell_2$-regularization, so a~priori is not guaranteed if IFs that rely on Hessian approximations would fair better than the gradient heuristics, \randproj and \tracin.
On the test set \emph{newstest16} 
we obtained BLEU 
36.0 after 100k training steps. 

To construct a synthetic noisy dataset, we uniformly sampled 4096 examples from the training data and,
for 256 examples of those, randomly shuffle the target sides, and repeat the above to obtain 5 data folds. We then apply different methods to compute self-influence scores and evaluate their quality with AUC and AP, averaging over the~5 data folds. 

From Table~\ref{tab:retrieve-mislabel-wmt17} we observe that \randselect performs the worst in terms
of retrieval quality. \arnoldi outperforms \randproj and we observe that here AP is a measure more sensitive to differences than AUC. The memory footprint of \randselect scales poorly: while we managed to run \arnoldi  with $b=512$ to compute self-influence, for \randselect we had to reduce it to 64. 

Finally, we consider the question of the quality of influence estimates obtained for subsets of layers. For
large models it is common to reduce $p$ by looking at the last few layers~\cite{tracin, fast_if, han-etal-2020-explaining}. However, \citet{feldman_practical} observed for their estimator $\mathcal{I}_{\rm mem}$ a degradation of the self-influence estimates computed using only the last layer of ResNet50, conjecturing that memorization of difficult examples has already happened by the time the computation reaches the last layer. We corroborate their findings here for Transformers: using only the last three decoder layers we observe a significant degradation of the retrieval quality for all algorithms, with \arnoldi still outperforming the rest.
\begin{table}
    \centering
    \resizebox{0.9\columnwidth}{!}{
    \begin{tabular}{llrrr}
    \toprule
         Layers & Method & $\tilde{p}$ & AUC & AP \\
         \midrule
         \multirow{6}{*}{\makecell{all \\ layers}}&\randselect & 10 & 67.0 & 12.2 \\
         &\randselect & 100 & 79.3 & 19.7 \\
         \cline{2-5}
         &\randproj & 10 & 85.6 & 31.3 \\
         &\randproj & 20 & 85.2 & 28.0 \\
         \cline{2-5}
         &\arnoldi & 10 & 92.0 & 47.8 \\
         &\arnoldi & 20 & \textbf{93.8} & \textbf{54.4} \\
         \midrule
         \midrule
         \multirow{6}{*}{\makecell{last 3 \\ decoder \\ layers}}&\randselect & 100 & 80.2 & 20.1 \\
         &\randselect & 1000 & 81.3 & 22.2 \\
         \cline{2-5}
         &\randproj & 10 & 80.6 & 23.5 \\
         &\randproj & 20 & 83.0 & 25.8 \\
         \cline{2-5}
         &\arnoldi & 10 & 82.0 & 28.1 \\
         &\arnoldi & 20 & \textbf{83.7} & \textbf{28.5} \\
         \bottomrule
    \end{tabular}}
    \caption{Retrieving synthetic mislabeled examples on WMT17. The standard deviation of AUC and AP estimates were under, resp., 0.9 and 1.0. for the \texttt{Rand*} methods.}
    \label{tab:retrieve-mislabel-wmt17}

\end{table}

\paragraph{Filtering noisy training corpus.}
To simulate a realistic data cleaning setup, we take the noisy Paracrawl corpus from the WMT18 Parallel Corpus Filtering Task\footnote{\url{http://statmt.org/wmt18/parallel-corpus-filtering.html}}. This dataset consists of 100M German-English sentence pairs with different kinds of noise that naturally occur in Web crawled datasets: sentence pairs in languages others than English or German; where both sides are in English or German; where the target is either not or only a partial source translation, or is a near copy of the source; or non-informative pairs, e.g.~short sentences consisting of numbers. 

We trained a model on WMT17 to bootstrap self-influence calculation and used \emph{newstest2016} for evaluation of retrained models. The methods that scale to this data size are \arnoldi, \randproj and \randselect, but as the latter underperformed on the retrieval of synthetically mislabeled data and has heavy memory consumption, we focused on the former two. We chose $\tilde{p}=10$ as it is faster than $\tilde{p}=20$, which did not substantially increase the scores in Table~\ref{tab:retrieve-mislabel-wmt17}. With this, self-influence scoring speed was 2.2M examples/hour on a 16-core TPUv2 using $b=2048$. 

As filtering baselines we consider training on the full uncleaned data and some pre-filtering strategies from the AFRL, Alibaba and Microsoft submissions~\cite{afrl_submission, alibaba_submission, microsoft_submission} that are detailed in \S\ref{app:mt_filtering}. 
%
In Table~\ref{tab:noisy-Paracrawl-selection} we report results after training on 1\%, 5\% and 10\% of the data with the lowest self-influence, i.e.~the cleanest data by assumption.  We followed~\cite{vaswani2017all} and reported BLEU scores both at 10k steps and at the final 200k steps as the gap between data selection strategies might reduce over the course of training,
possibly as gradient updates from cleaner data might be favored over time. 
Both \arnoldi and \randproj select cleaner data at the bottom of the 5\% and 10\% self-influence scores. Also \arnoldi outperforms \randproj gaining almost 4 BLEU points at 10k steps, and more than 8 points over the pre-filtering baseline at 200k steps.
At 1\%, we see a degradation in the performance at 200k steps and, inspecting the training loss, we found that it decreases by more than a half going from 5\% to 1\% data threshold. We conjecture that the selected examples at this strict 1\%-threshold are too ``simplistic'' to provide useful translation patterns, in line with similar findings of~\citet[\S 3.3]{feldman_practical} on the marginal utility of data with low memorization values on ImageNet. See \S\ref{app:mt_exemplars} for examples of high/low influence sentences.

Above, we did not aim to beat state-of-the-art cleaning pipelines, which employ a cascade of filtering stages, but to investigate whether a simple application of IFs can select better data to train on. Nevertheless, we evaluated the added value of \arnoldi by filtering 25\% and 50\% of the \emph{clean} data selected by Microsoft's cascade (WMT18 winner), and this increased BLEU from their 36.32 to, resp., 37.38 and 37.20 on \emph{newstest16}.

\begin{table}
    \centering
    \resizebox{0.9\columnwidth}{!}{
    \begin{tabular}{lrrr}
    \toprule
    Method & \% selected & BLEU@10k & BLEU@200k \\
    \midrule
    None & 100 & 9.9 & 17.8 \\
    Pre-filtering & 14 & 11.7 & 22.6 \\
    \midrule
    \midrule    
    \randproj & 1 & 7.7 & 8.7 \\
    \arnoldi & 1 & 17.8 & 19.6 \\
    \midrule
    \randproj & 5 & 18.7 & 27.9 \\
    \arnoldi & 5 & 24.0 & 30.3 \\
    \midrule
    \randproj & 10 & 21.3 & 28.6 \\
    \arnoldi & 10 & \textbf{25.0} & \textbf{30.8} \\
    \bottomrule
    \end{tabular}}
    \caption{Data selection on the noisy Paracrawl corpus (100M parallel sentences) with evaluation on \emph{newstest16}.}
    \label{tab:noisy-Paracrawl-selection}
\end{table}
\subsection{Scaling with Model Size: Computer Vision}
Here we empirically verify if \arnoldi scales well with the number of parameters for four state-of-the-art computer vision models of increasing sizes, and also run the mislabeled data retrieval experiment for the largest Vision Transformer.

\paragraph{Method performance.} We considered ResNet50 and Vision Transformers (ViT): the Flax implementation\footnote{https://github.com/google/flax/tree/main/examples/imagenet} of ResNet50 has about 25M parameters\footnote{Even though the model parameter count is about 50M, half of the parameters are batch statistics, so we treat $p$ as 25M.}, and for ViTs we used the JaX implementation\footnote{https://github.com/google-research/vision\_transformer} and took checkpoints trained on the mixture of ImageNet and ImageNet21k, that have between 86M (B32) and 300M (L32) parameters. Additionally, to cover an intermediate size between the ViT B32 and L32 we took a subset of the layers, starting from the top, to compute influence w.r.t.~50\% of the parameters of the ViT L32, amounting to 150M of weights. For all models we used one 4-core TPUv2, $n=200$ and $b=512$ (see \S\ref{app:vit_details}).

\begin{figure}
\centering
    \includegraphics[width=0.9\columnwidth]{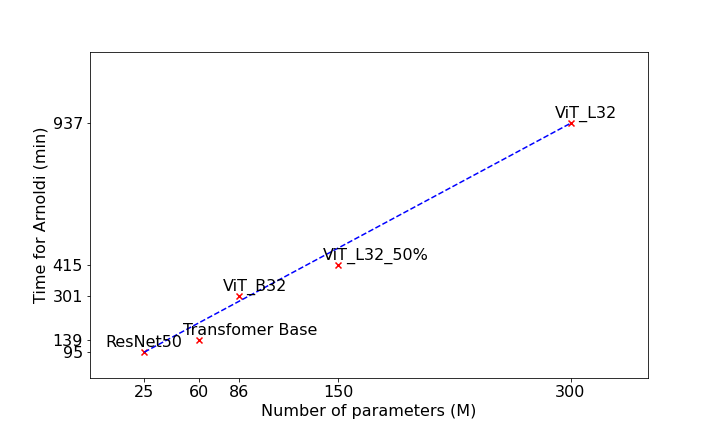}
    \caption{Runtime of \arnoldi for $n=200$ iterations on the full set of parameters of respective networks.}
    \label{fig:arnoldi timings}
\end{figure}

Figure~\ref{fig:arnoldi timings} plots the time taken by an \arnoldi run on the full set of parameters of the considered models (for a better perspective, we also included the translation Transformer Base model). As expected, we observe a linear trend in terms of model size, with the largest-size model (ViT-L32) taking 15 hours to estimate top-200 (cachable) eigenvalues.

\paragraph{Retrieving mislabeled data after fine-tuning.}
As vanilla ViTs have been reported to have extremely sharp loss landscapes~\cite{vit_landscape}, we took their ViT L32 checkpoint trained with the landscape-smoothing SAM loss~\cite{sam} to better satisfy IF assumptions, and \emph{fine-tuned} it on CIFAR10 for 10k steps obtaining a 98.6\% test accuracy.
Since fine-tuning converged to a saddle point\footnote{Among the top-100 $\lambda$s, 7\% of the mass belongs to negative $\lambda$s.}, we restricted the \mbox{\textsc{Influence}} procedure only to $\lambda_i>0$. 

We mislabeled 10\% of the test examples and compared their retrieval by~\arnoldi and~\randproj accumulating parameters from the top 10\%, 20\% and the full model in \S\ref{app:mis_vit}, Table~\ref{tab:cifar10}. \arnoldi wins, but the gap to \randproj is small and perhaps indicates that accounting for local curvature is superfluous for this particular self-influence benchmark and model. As demonstrated by~\citet[\S 2.2]{koh2017understanding}, this may not be the case in general, and for other IF-based applications our contribution enables a verification of whether accounting for Hessians is required. Unlike for the machine translation case, increasing the number of parameters leads here to a slight decrease in performance, suggesting that one may restrict IFs to the top layers in the fine-tuning setting.
Another reason 
for the near matching performance 
could be 
the IF's increased fragility for large models~\cite{basu}, also called out for natural language inference and RoBERTa model by~\citet{kocijan20}, where performance dropped after retraining on either high- or low-influence (w.r.t.~to a validation set) examples. In~\S\ref{app:mem_gen_tradeoff} we also investigate the memorization vs.~generalization trade-off of removing high or low self-influence images on ImageNet.

In~\S\ref{app:resnet}, for the whole ImageNet and the full ResNet50 model, we picture most self-influential images and the most influential training images retrieved for a test point. 



\section{Conclusion}
We proposed a new way of calculating influence scores of~\cite{koh2017understanding} for large DNNs by approximate diagonalization of their Hessians and avoiding re-estimating them on every training example.
We demonstrated finding influential or noisy examples in datasets of up to 100M training examples and models with up to 300M parameters.

As we showed, depending on task's nature, its convergence status and local convexity, IFs can be either superior or match random projection and gradient-based approaches, when measured on the benchmark of retrieving synthetically mislabeled data with self-influence scores. Whether this holds for other influence-based tasks, like understanding learned patterns~\cite{han-etal-2020-explaining}, debugging data biases~\cite{brunet2019understanding}, or estimating data quality~\cite{wang2018data}, requires further investigation. Importantly, we provided an efficient tool to enable such analyses within acceptable timeframes on standard hardware for models and datasets that matter in applications.



\section*{Acknowledgements}
We thank Behrooz Ghorbani and Mukund Sundararajan for their valuable feedback on the paper.

{\small\bibliography{main.bib}}

\appendix
\begin{figure}[H]
\centering
\includegraphics[width=0.6\columnwidth]{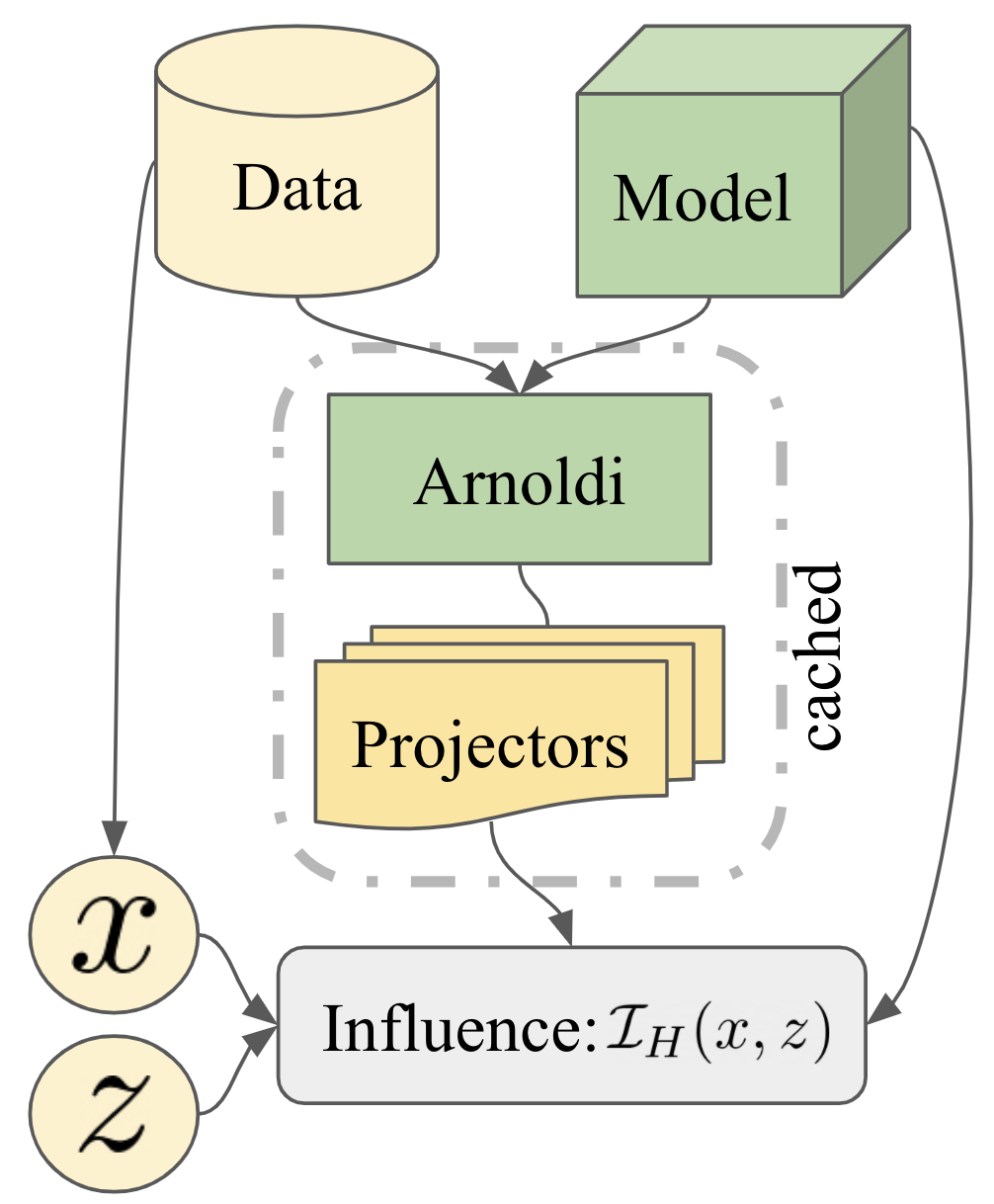}
\caption{General illustration of our approach.}
\label{fig:res-diagram}
\end{figure}

\section{Approximation of Influence Functions}\label{app:approximation}
In this section we discuss the derivation of influence functions and the
effect of restricting the Hessian to the top-$\tilde p$ eigenvalues on the estimates.
We consider a point $x$ to be removed from the training data $\traindata$ of
cardinality $N$. We let $L(x|\Theta)$ denote the loss at $x$ under the parameters $\Theta$
and $L(\traindata|\Theta)$ the total loss:
$$
L(\traindata|\Theta) = \frac{1}{N}\sum_{x\in \traindata}L(x|\Theta).
$$
In this section we make the dependence of the Hessian $H$ on the learned parameters $\Theta_0$
explicit by writing $H=H_{\Theta_0}$.
\begin{lemma}\label{lmm:q}
Let $\Theta_0$ be the parameters at convergence and assume that $\nabla_\Theta L(\traindata|\Theta_0)=0$.
Then up to terms of order $o(1/N^2)$, the effect of retraining on $\traindata\setminus\{x\}$ is approximated
by shifting the parameters by $\theta_*$ which minimizes the quadratic form
$$
Q(\theta) = \frac{1}{2}\langle\theta, H_{\Theta_0}\theta\rangle -\frac{1}{N}\langle\nabla_\Theta L(x|\Theta_0),\theta\rangle.
$$
In particular, up to a constant factor, the influence of $x$ on $z$ is approximated by
\begin{equation}\label{eq:supp_hessian_influence}
\mathcal{I}_H(x, z) = \langle \nabla_{\Theta_0}L(x), H_{\Theta_0}^{-1} \nabla_{\Theta_0}L(z)\rangle.
\end{equation}
\end{lemma}
\begin{proof}
We follow the argument in~\cite{koh2017understanding} and seek to model retraining as minimization of:
$$
f(\theta) = L(\traindata|\Theta_0+\theta) - \frac{1}{N}L(x|\Theta_0+\theta);
$$
we then assume that $\|\theta\|_2$ is $O(1/N)$, do a Taylor expansion and drop terms of order $O(1/N^2)$ obtaining:
\begin{equation*}
\begin{split}
f(\theta) &\approx L(\traindata|\Theta_0) - \frac{1}{N}L(x|\Theta_0) -\frac{1}{N}\langle\nabla_\Theta L(x|\Theta_0),\theta\rangle \\
&+ \frac{1}{2}\langle\theta, H_{\Theta_0}\theta\rangle,
\end{split}
\end{equation*} which is $Q(\theta)$ up to a constant. The minimizer of $Q$ is then
$$
\theta_* = \frac{1}{N}H_{\Theta_0}^{-1}\nabla_\Theta L(x|\Theta_0),
$$ and the first order Taylor expansion of $L(z|\Theta_0+\theta)-L(z|\Theta_0)$ yields~\eqref{eq:supp_hessian_influence} up
to the multiplicative factor $1/N$.
\end{proof}
Let us now order the eigenvalues of $H_{\Theta_0}$ in non-decreasing absolute value:
$
|\lambda_1| \ge |\lambda_2| \ge \cdots \ge |\lambda_k| \ge \cdots
$
and denote by $e_1, e_2, \cdots, e_k,\cdots,$ the corresponding eigenvectors. In~\cite[Sec.~4 and~E]{spectrum_guys} it is empirically observed (and proven in the quadratic case) that during training dynamics the gradient updates
will tend to align with the subspace of $H_{\Theta_0}$ corresponding to the dominant eigenvalues. Here
we provide two different arguments in the same direction. First, we upper bound the error of approximating
$Q$ using such a subspace; we then discuss the effect of noise depending
on the size of the eigenvalues when applying the inverse Hessian to a vector. Note that here we
are under the idealized conditions of the influence functions approximations~\cite{koh2017understanding} where the loss
is well-approximated by a quadratic form.

If we restrict $H_{\Theta_0}$ to the subspace spanned by the top-$k$ eigenvalues we then obtain a new quadratic form
$Q_k$. Intuitively, as $k$ increases $Q_k$ better approximates $Q$; moreover, we expect that errors in the estimation of
$Q(\theta)$ in directions of the eigenvectors corresponding to smaller eigenvalues should matter less.
\setcounter{lemma}{0}
\begin{lemma}
The form $Q_k$ approximates $Q$ by an error bounded by:
$$
0\le Q(\theta) - Q_k(\theta) \le  \frac{1}{2}|\lambda_{k+1}| \|\theta\|_2^2.
$$
If minimizing $Q$ introduces an error $\varepsilon$ in the direction of $e_{k+1}$ obtaining an estimate $\theta'$ for $\theta_*$, then 
$$
Q(\theta')-Q(\theta_*)=\frac{\varepsilon^2}{2}\lambda_{k+1}.
$$
\end{lemma}
\begin{proof}
If we decompose $H_{\Theta_0}$ using the eigenvectors we get:
$$
\langle \theta, H_{\Theta_0}\theta\rangle = \sum_{j=1}^d \lambda_j^2 (\langle\theta, e_k\rangle)^2.
$$
However, when we build $Q_k$ we only consider terms up to $j=k$ so
$$
Q(\theta) - Q_k(\theta) = \frac{1}{2} \sum_{j=k+1}^d \lambda_j^2 (\langle\theta, e_k\rangle)^2\le \frac{1}{2}|\lambda_{k+1}| \|\theta\|_2^2.
$$
For the second part we have $\theta'=\theta_*+\varepsilon e_{k+1}$; using the symmetry of $H_{\Theta_0}^{-1}$ we obtain the 
following expansion:
\begin{equation*}
\begin{split}
    Q(\theta_* + \varepsilon e_{k+1}) &= Q(\theta_*) -\frac{1}{N}\langle \nabla_\Theta L(x|\Theta_0), \varepsilon e_{k+1}\rangle \\
        &+\frac{1}{2N}\langle \nabla_\Theta L(x|\Theta_0), \varepsilon e_{k+1}\rangle \\
        &+\frac{1}{2N}\langle H_{\Theta_0}^{-1}\nabla_\Theta L(x|\Theta_0),\varepsilon  H_{\Theta_0} e_{k+1}\rangle \\
        & + \frac{1}{2}\langle \varepsilon e_{k+1}, \varepsilon  H_{\Theta_0} e_{k+1}\rangle 
        = \frac{\varepsilon^2}{2}\lambda_{k+1}.
\end{split}
\end{equation*}
\end{proof}

Below we look at the same
phenomenon from a theoretical perspective by showing that the variance of estimating influence increases if one
looks at the small eigenvalues of $H_{\Theta_0}$. To formalize, assume that we want to compute
$$
y = H_{\Theta_0}^{-1}u
$$
and that $u$ can be estimated exactly. We also assume that $H_{\Theta_0}$'s estimation is noisy resulting in
$H_{\Theta_0}+\delta H$; and that the expected value of $\delta H$ is $0$ and that the $\delta H$
is isotropic (e.g.~it does not preferentially align with some $e_k$ or co-vary with $\lambda_k$). We then show that for the estimator
$$
\hat y = (H_{\Theta_0}+\delta H)^{-1}u
$$
the variance in the direction of $e_k$ is proportional to $|\lambda_k|^{-2}$; this formalizes the intuition that 
the small eigenvalues increase the noise in the estimates.
\begin{lemma}
The variance of $\hat y$ in the direction of $e_k$ is:
$$
{\rm Var}(\langle\hat y, e_k\rangle) \approx \frac{1}{|\lambda_k|^2}{\rm Var}(\langle \delta H e_k, y\rangle).
$$
\end{lemma}
\begin{proof}
The estimator $\hat y$ solves another problem than $y$:
$$
(H_{\Theta_0} + \delta H)\hat y = u = H_{\Theta_0}y;
$$ if we write $\hat y = y + \delta y$ and ignore the term $\delta H \cdot \delta y$ we obtain:
$$
\delta y = - H_{\Theta_0}^{-1}\delta H \cdot y,
$$
from which
$$
\langle \delta y, e_k\rangle = -\frac{1}{\lambda_k}\langle e_k, \delta H \cdot y\rangle,
$$
from which the result follows using that $\delta H$ is symmetric.
Note that from the above equation we can also deduce, having ignored the  term $\delta H \cdot \delta y$,
that $\mathbb{E}[\delta y] \approx 0$, so that $\hat y$ is approximately an unbiased estimator of $y$ too.
\end{proof}
\subsection{Arnoldi iteration for inverse Hessian}\label{app:inverse_arnoldi}
Note that in the ideal situation, in which \arnoldi correctly locates the top-$k$ eigenvalues and eigenvectors, one is minimizing $Q_k$ instead of $Q$. On the other hand, one might try to apply
\arnoldi directly to $H_{\Theta_0}^{-1}$ obtaining an approximation in operator norm using the
subspace corresponding to the top-$k$ eigenvalues of $H_{\Theta_0}^{-1}$, which correspond to the $k$
eigenvalues of $H_{\Theta_0}$ with smallest absolute value. However, this approach also requires estimating inverse HVPs iteratively, e.g.~using \lissa. For example, if \lissa is employed with $r$ iterations for each iteration of \arnoldi, this approach is $r$ times slower. We also did not find this approach competitive: for the larger CNN network considered in~\S\ref{sec:mnist} we tested it using $r=50$ and, while for $\tilde p=10$ we got a slightly better retrieval quality of 97.9/91.9 (AUC/AP), with $\tilde p=100$ the retrieval quality was not better than $\arnoldi$'s 97.9/92.8.

\section{Corrupted MNIST}\label{app:mnist}
\subsection{Opponents and Proponents}\label{app:mnist_oppo}
Following~\cite{tracin}, we will call, for a fixed $z$, those $x$ where $\mathcal{I}_H(x,z) > 0$  \emph{proponents} of $z$, as removing them from the training data increases the loss at $z$; and those $x$ where $\mathcal{I}_H(x,z) < 0$,
\emph{opponents} of $z$ as their removal decreases the loss at $z$. 

We use the CNN network with 800k parameters from~\S\ref{sec:mnist} and $\tilde p=10$ Arnoldi projectors
to find top opponents and proponents of test images. Note that, to highlight the fact
that IFs without normalization tend to retrieve examples with bigger gradients, we compute proponents and opponents both with and without normalizing all gradients to a unit norm. Scaling influence scores by (transformed) gradient norm has already been used to compensate for IF's propensity to prefer high-loss examples~\cite{relatif}, while not using normalization may have lead to negative conclusions about IF usefulness~\cite{kocijan20}.



\begin{figure*}
\centering
\includegraphics[width=\textwidth]{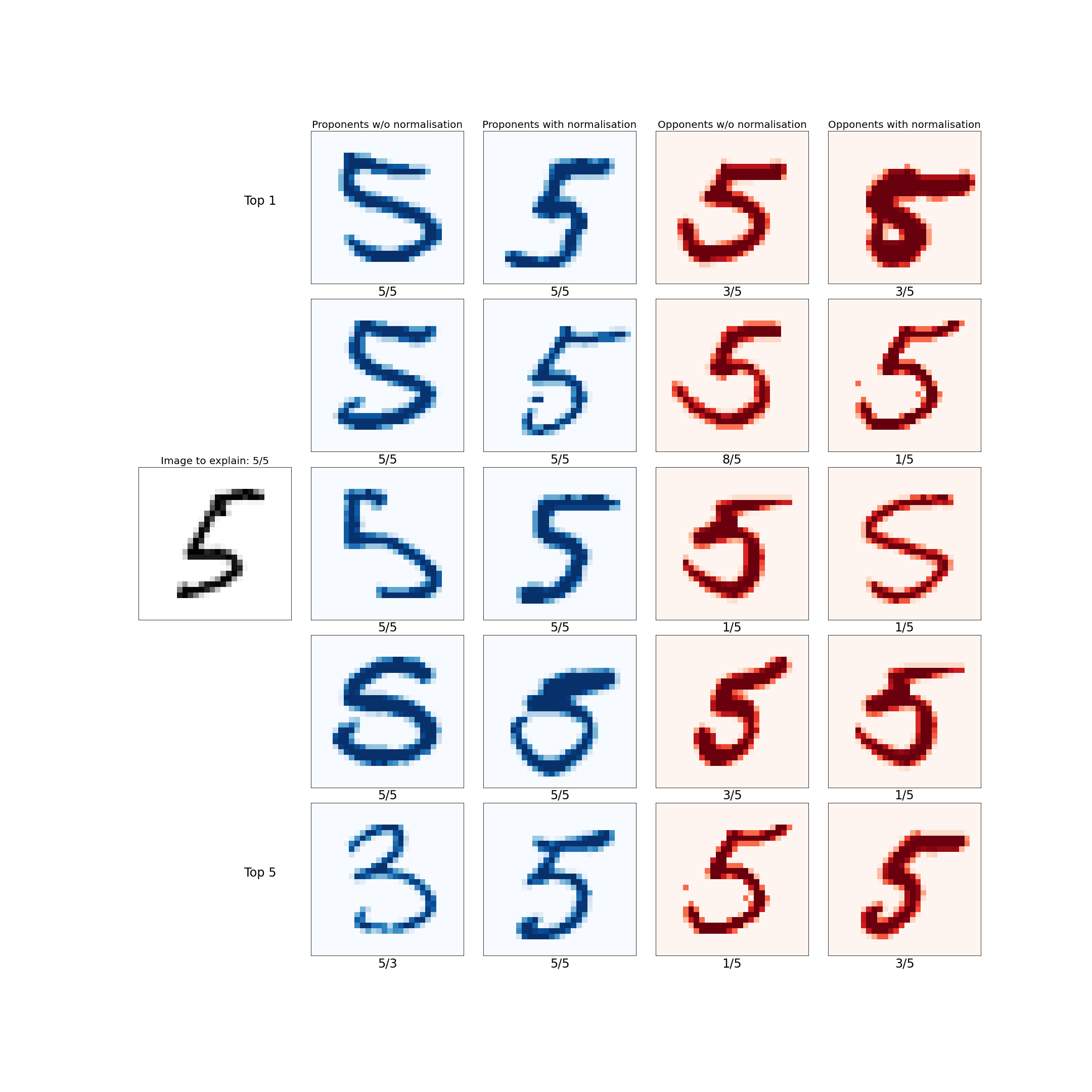}
\caption{Proponents and opponents of correctly labeled test point.
Proponents are colored in blue and opponents in red while the test point is in gray at the center. For each point
we report the label in the corrupted MNIST / the label in the original MNIST.}
\label{fig:mnist-proponents-and-opponents-5}
\end{figure*}

Overall, we find that results are easily interpretable by a human. 
Let us first consider the example
in Figure~\ref{fig:mnist-proponents-and-opponents-5} for a correctly labeled digit 5.
Except for one case, proponents are correctly labeled 5s and using gradient normalization
retrieves at the top 3, examples of 5s which are stylistically similar to the original image.
On the other hand, without gradient normalization, a strong proponent can also be
a fake~5, because the gradients can be big. In this case we have a 3 labeled as 5 as the 5th proponent.
Opponents are all instances of 5s that were incorrectly labeled.

\begin{figure*}
\centering
\includegraphics[width=\textwidth]{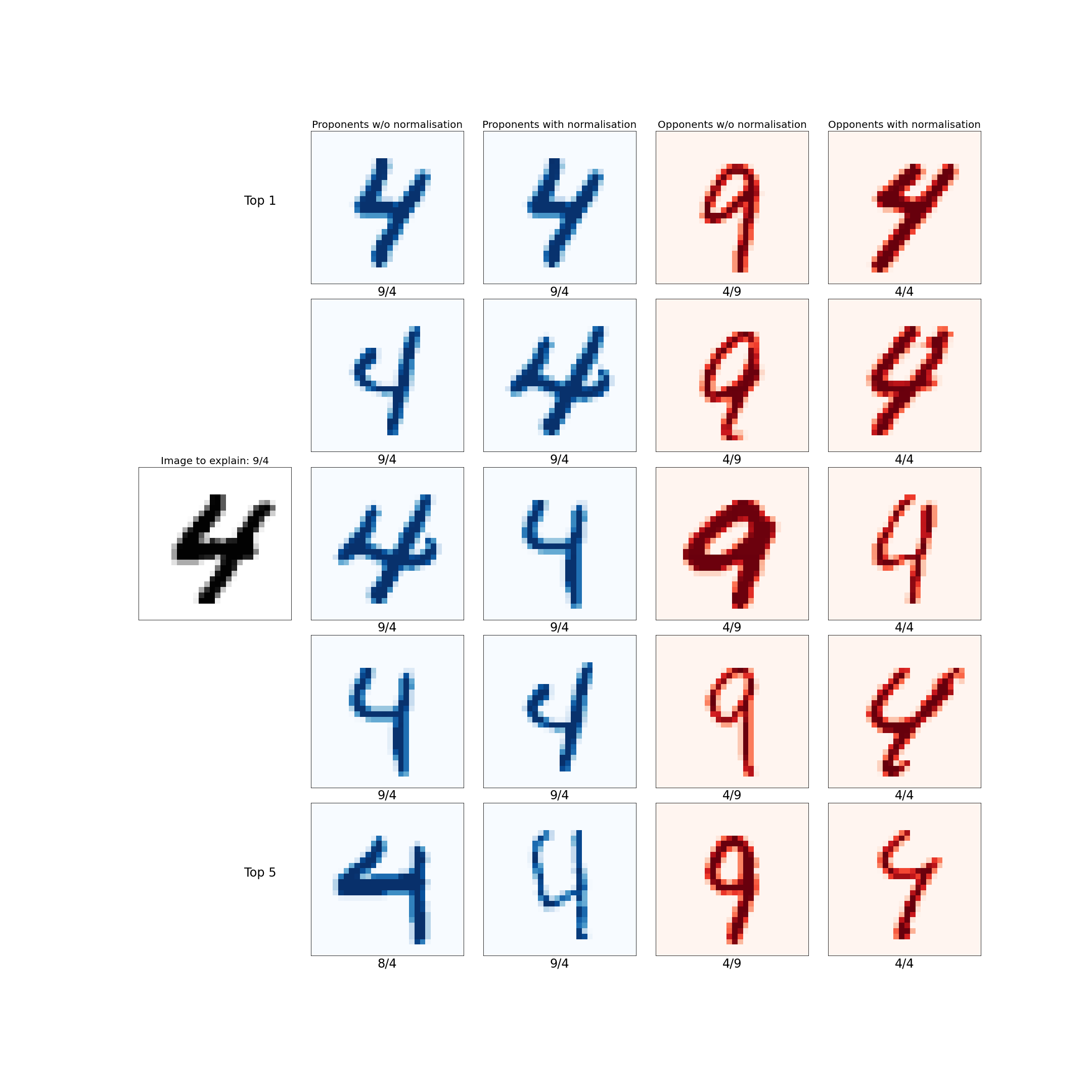}
\caption{Proponents and opponents of an incorrectly labeled test point.
Proponents are colored in blue and opponents in red while the test point is in gray at the center. For each point
we report the label in the corrupted MNIST / the label in the original MNIST.}
\label{fig:mnist-proponents-and-opponents-4}
\end{figure*}

Now consider digit 4 mislabeled as 9 in Figure~\ref{fig:mnist-proponents-and-opponents-4}.
All the proponents are 4s incorrectly labeled as 9. Only 
without gradient normalization the top-5th proponent is a 4 labeled as an 8, but one might also
argue that it looks close by a 9 (if one would add a small closing stroke at the top). Opponents
without normalization are instances of 4s correctly labeled as 4; while opponents without normalization
are 9s labeled as 4; this is expected as without normalization instances with bigger gradients
will be preferred.

\begin{table}[t]
    \centering
   \resizebox{0.8\columnwidth}{!}{
    \begin{tabular}{lrrrrr}
\toprule
  Method &    $\tilde{p}$  & AUC & AP & AUC & AP\\
\toprule
&&\multicolumn{2}{c}{w/o $||\Theta||_2^2$ }& \multicolumn{2}{c}{with  $||\Theta||_2^2$ }\\
\midrule
\lissa, $r=10$ & - & 76.6  & 43.6 & 96.6 &87.3\\
\midrule
\tracin[1] & -  & 63.1 & 39.9 & 99.3 & 97.9 \\
\tracin[10] & -  & \textbf{92.3} & \textbf{72.1} & \textbf{99.7} & \textbf{99.0} \\
\midrule
\randproj & 10  &76.4  & 43.5 & 91.5 & 72.8\\
\randproj & 100 &76.6 &43.6 & 95.1 & 83.4\\
\midrule
\randselect & 10  & 56.4 & 28.9 &73.3  &50.8  \\
\randselect & 100 & 63.0 & 36.2 &97.1  & 81.5 \\
\midrule
\arnoldi & 10  & 74.7&39.5 & 86.7 & 71.3 \\
\arnoldi & 100 & 75.5 &42.3 & 92.3 & 80.2 \\
\bottomrule
\end{tabular}}
    \caption{Retrieval of mislabeled MNIST examples using self-influence for the longer trained large CNN model, with and without $\ell_2$-regularization. For \tracin, the
    number of checkpoints is in brackets (one last or all). All methods use full models. Scoring time $T$ is omitted since it exactly matches the numbers in Table~\ref{tab:mislabel-mnist-retrieval}.}
    \label{tab:mislabel-mnist-retrieval_longer_cnn}
\end{table}

\subsection{Longer trained larger network}\label{app:longer_cnn}
To investigate how longer training effects the methods' performance on the CNN network with 800k parameters from~\S\ref{sec:mnist}, we trained it for 100 epochs instead of 10, both with and without $\ell_2$-regularization (denoted as $||\Theta||_2^2$ in Table~\ref{tab:mislabel-mnist-retrieval_longer_cnn}). Without regularization, CNN accuracy on the test set is 69.47\% which points to the model over-fitting and leads to much lower AUC/AP for all the methods; regularization prevents over-fitting and lifts the accuracy to 74.3\%. Despite different training protocols, we observe the same ranking of methods as in Table~\ref{tab:mislabel-mnist-retrieval} in~\S\ref{sec:mnist}. As before, \tracin on 10 checkpoints has the best retrieval accuracy for this CNN, but it is much slower than other methods except for \lissa, and does not scale to even larger models and datasets. The poor performance of \tracin on the last checkpoint without regularization, highlights the danger of replacing influence with gradient similarity for models at risk of over-fitting. The projection-based \arnoldi and \randproj however, given enough projectors, fair better in this setup.

\section{Machine Translation}
\subsection{Prefiltering Baseline}\label{app:mt_filtering}
For machine translation we implemented a prefiltering baseline inspired by the AFRL, Alibaba and 
Microsoft submissions~\cite{afrl_submission, alibaba_submission, microsoft_submission} to the WMT18 corpus filtering task:
\begin{itemize}
    \item \textbf{AFRL}: remove pairs where the ratio of the number of English to German words is above 3 or
    below 1/3. This is likely to eliminate misaligned pairs. We also
    filtered out pairs where at least one side contains a word consisting of more than 30 tokens as a heuristic to remove long URLs.
    \item \textbf{Alibaba}: remove pairs where the edit distance between the two segments is less than 2 or the relative
    edit distance (i.e.~normalized by the average of the segments' lengths) is less than 0.1. This removes segments
    where the target is a minor modification of the source.
    \item \textbf{Microsoft}: use a language classifier to remove segments where either the target is not English
    or the source is not German. As the Microsoft submission did not release their classifier,
    we used the one from~\cite{zhang-etal-2018-fast}.
\end{itemize}

\subsection{Experimental Details}\label{app:mt_details}

We used the Flax library\footnote{https://github.com/google/flax/tree/main/examples/wmt} and the 
configuration for Transformer Base. Specifically, we used a shared SentencePiece~\cite{kudo2018sentencepiece} vocabulary of 32k tokens,
hidden dimension of 512, MLP dimension 2048, 8 attention heads and 6
layers in the encoder and the decoder. Note that we share weights between the last layer and the embeddings, and use bfloat16 precision. The dropout rate was 0.1 both for
attention and MLPs. 

To improve TPU utilization, and effectively increasing the number of tokens in each batch, we packed sentences~\cite{shazeer2018mesh} with a max sentence length 256; this results
in about 56k non-padding tokens in each batch. We used Adam with the defaults of the Flax library. For the learning schedule we increase linearly the learning rate to 0.002 in the first
1k steps and then use square-root decay.
For BLEU evaluation we used SacreBLEU\footnote{https://github.com/mjpost/sacrebleu; configuration:\\
{  case:mixed+eff:no+tok:13a+smooth:exp+version:2.0.0}}.

\newcommand{\wrapexemplar}[1]{\begin{minipage}[t]{16.5cm}#1\end{minipage}}

\subsection{Exemplars}\label{app:mt_exemplars}
In Table~\ref{tab:mt-exemplars} we report some exemplars of data classified as having lower or higher self-influence by \arnoldi on the noisy Paracrawl. In general, up to the 1\% quantile the data consists of clean but very short segments; between 1\% and 10\% the data
becomes more interesting with longer translated segments, that may not always be exact translations though. The exemplar in the 90\%-100\% shows an incorrect pair of two English sentences which, being highly-self influential, requires memorization from the model.

To better understand the cleaner portion of the data we looked at the Spearman correlation between length of the
longest sentence (German or English) and the 
quantile rank. For the data in the 0-10\% quantile we observe a highly significant correlation of about 0.5 with a $pval < 10^{-6}$, justifying popular short-segment removing heuristics.
On the other hand, this correlation disappears as the data gets less clean; for example, in the range 70\%-100\% the 
Spearman correlation is negative and of just -0.19.

\begin{table*}[h]
\centering
\resizebox{\textwidth}{!}{
     \begin{tabular}{lll}
   \toprule
   &Quantiles &    Parallel Exemplar \\
   \midrule
   & [0\%, 1\%] & \wrapexemplar{DE: Statistik Schweiz ~~~ EN: Statistics Switzerland} \\
   \midrule
  & [1\%, 5\%] & \wrapexemplar{DE: 1. Darf ich dieses Buch kaufen? ~~~ EN: 1. May I buy this book ?} \\
   \midrule
   & [5\%, 10\%] & \wrapexemplar{DE: Die bis zu 15.000 Bogen/h schnelle, gegenüber vielen anderen Modellen in dieser Formatklasse besonders platz- und energiesparende Maschine wird mehrmals täglich bei dem Druck und Inline-Veredelung von hochwertigen Postern live präsentiert \\
EN: The press is engineered for production speeds up to 15,000 sheets per hour and boasts not only a significantly smaller footprint, but also greatly reduced energy consumption compared to many other presses in this format class.} \\
  \midrule
& [90\%, 100\%] & \wrapexemplar{DE: View Public Profile ~~~
  EN: The time now is 01:18 pm .} \\
 \bottomrule
 \end{tabular}}
     \caption{Exemplars of Paracrawl data from low to high self-influence as found by \arnoldi. Cleaner, low influence, data is not always 
     useful, as it can consist, e.g., of very short segments. Mislabeled data has higher high self-influencebecause such needs to be memorized verbatim, as the model cannot ``interpolate'' the label from similar examples.}
     \label{tab:mt-exemplars}
\end{table*}

\section{Computer Vision}\label{app:cv}

\subsection{Experimental Details for ViT}\label{app:vit_details}

Since memory consumption also scales linearly with $p$, to avoid exhausting the TPU memory, we offloaded the results of each iteration to the host RAM. A further optimization that we leave for future work is to improve the memory footprint by exploring procedures using restarts, which consists in not choosing the starting vector $v$ randomly or in caching intermediate estimates. One approach might consist in doing $x$ iterations to estimate the eigenvectors of the top $x$-eigenvalues, save them to disk and restart the Arnoldi procedure using only the top eigenvector as $v$. Care must be taken to ensure that the restarts preserve, at least approximately, the orthogonality of the Arnoldi projectors, which might require modifications to Algorithm~\ref{alg:arnoldi}.

\subsection{Mislabeled Examples Retrieval for ViT}\label{app:mis_vit}
For the synthetic noisy dataset, we uniformly sampled 10\% of CIFAR10 and randomly flipped their labels. The results of retrieval with ViT on CIFAR10 are in Table~\ref{tab:cifar10}. We ran \arnoldi with effective batch size of 8192 on a TPUv2 with $c=4$ cores using a $\frac{256}{c}$-step gradient accumulation.
\begin{table}[]
    \centering
    \resizebox{0.8\columnwidth}{!}{
    \begin{tabular}{llrrr}
    \toprule
         Parameters & Method & $\tilde{p}$ & AUC & AP \\
         \midrule
         \multirow{4}{*}{\makecell{top 10\%}} & \arnoldi & 10 & \textbf{94.2} & \textbf{64.4} \\
         & \randproj & 10 & 93.9 & 62.3 \\
         \cline{2-5}
         & \arnoldi & 20 & 93.9 & 61.3 \\
         & \randproj & 20 & 94.0 & 62.4 \\
         \midrule
         \midrule
         \multirow{4}{*}{\makecell{top 20\%}} & \arnoldi & 10 & 94.0 & 62.5 \\
         & \randproj & 10 & 93.8 & 61.7 \\
         \cline{2-5}
         & \arnoldi & 20 & 93.6 & 59.4 \\
         & \randproj & 20 & 93.9 & 61.9 \\
         \midrule
         \midrule
          \multirow{4}{*}{\makecell{100\% \\}} & \arnoldi & 10 & 93.1 & 57.0 \\
         & \randproj & 10 & 93.2 & 59.0 \\
         \cline{2-5} 
         & \arnoldi & 20 & 93.3 & 57.4 \\
         & \randproj & 20 & 93.7 & 59.2 \\
         \bottomrule
    \end{tabular}}
    \caption{Retrieving synthetic mislabeled examples on CIFAR-10. The standard deviation of AUC and AP estimates is resp. 0.1 and 0.7 over 5 runs.}
    \label{tab:cifar10}

\end{table}

\subsection{Mislabeled Examples Retrieval and Exemplars for ResNet50}\label{app:resnet}
\begin{figure*}
\centering
\includegraphics[width=\textwidth]{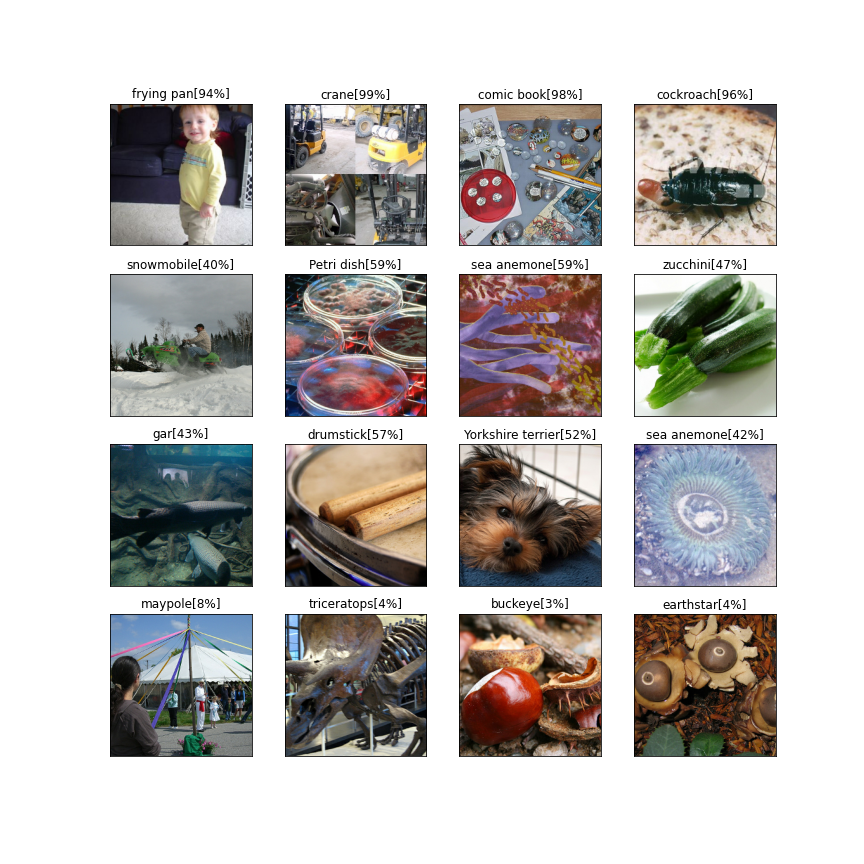}
\caption{Exemplars of ImageNet pictures scored by self-influence using ResNet50, with the corresponding quantile in brackets.}
\label{fig:res-net-exemplars}
\end{figure*}

In Figure~\ref{fig:res-net-exemplars}  we provide exemplars from ImageNet scored by self-influence using
RestNet50. We used the top-10 projectors extracted by \arnoldi. The top row shows random samples from the
top 90\%-100\% quantiles of self-influence scores, the two middle rows are from the 40\%-60\% quantile, and the bottom one represent the lower 0\%-10\% range. As can be seen, the top row contains either mislabeled images or ambiguous images with cluttered backgrounds containing more things than just the labeled object; in the mid-row some of the objects are in multiple counts and/or still have a rich background; in the last row, as a rule, the labeled object is prominently featured, occupying most of the image.

\begin{figure*}
\centering
\includegraphics[width=\textwidth]{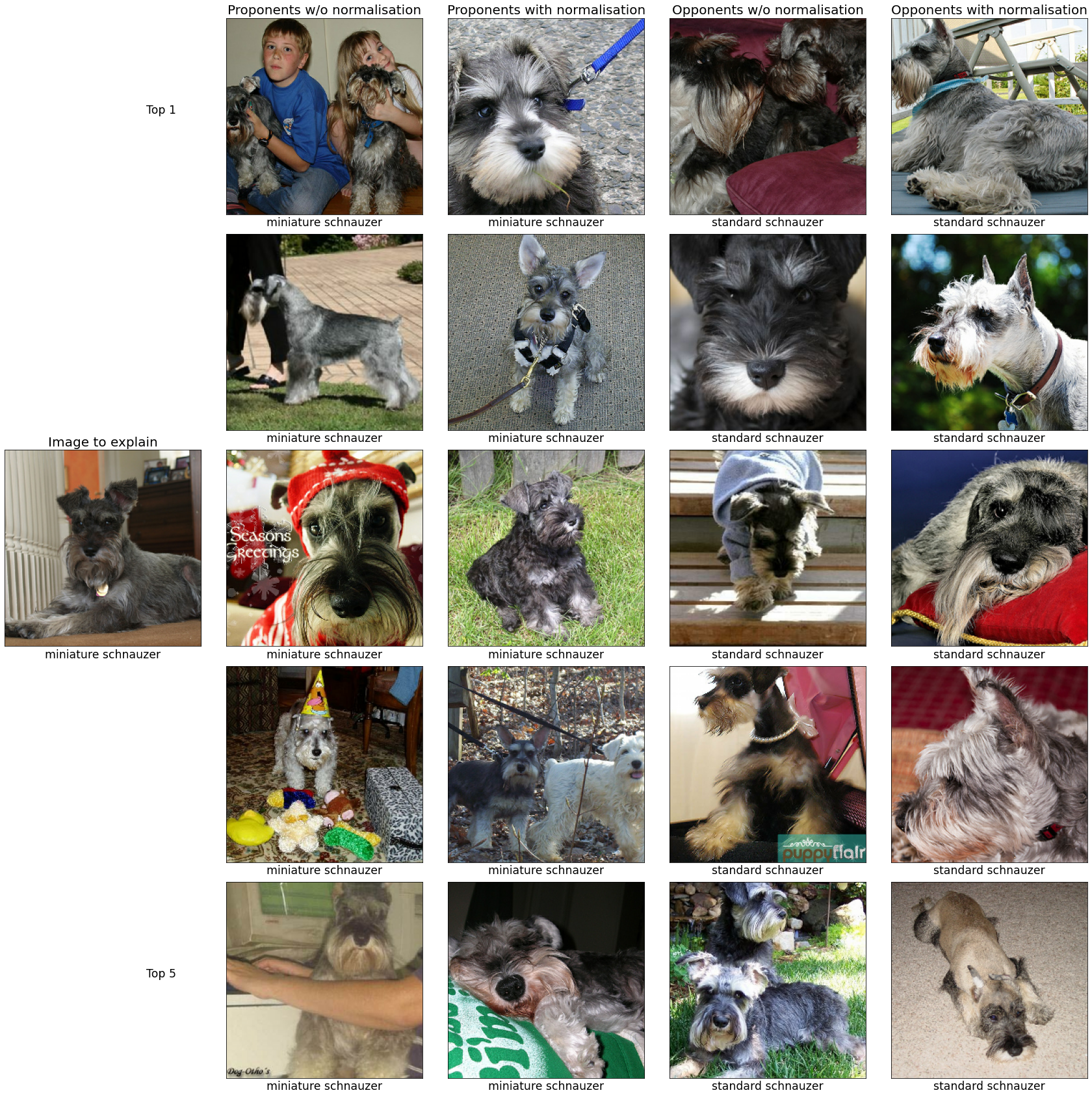}
\caption{Exemplar of ImageNet picture accompanied by its top 5 proponents and opponents with and without normalisation. We have pre-filtered projection based on the top 10 labels predicted by the model. Normalization of the gradients filters out the most "different" examples, for example, the top left image with children which had a large norm, was removed from the proponents after the normalization.}
\label{fig:res-net-exemplars_prop_op_1}
\end{figure*}

\begin{figure*}
\centering
\includegraphics[width=\textwidth]{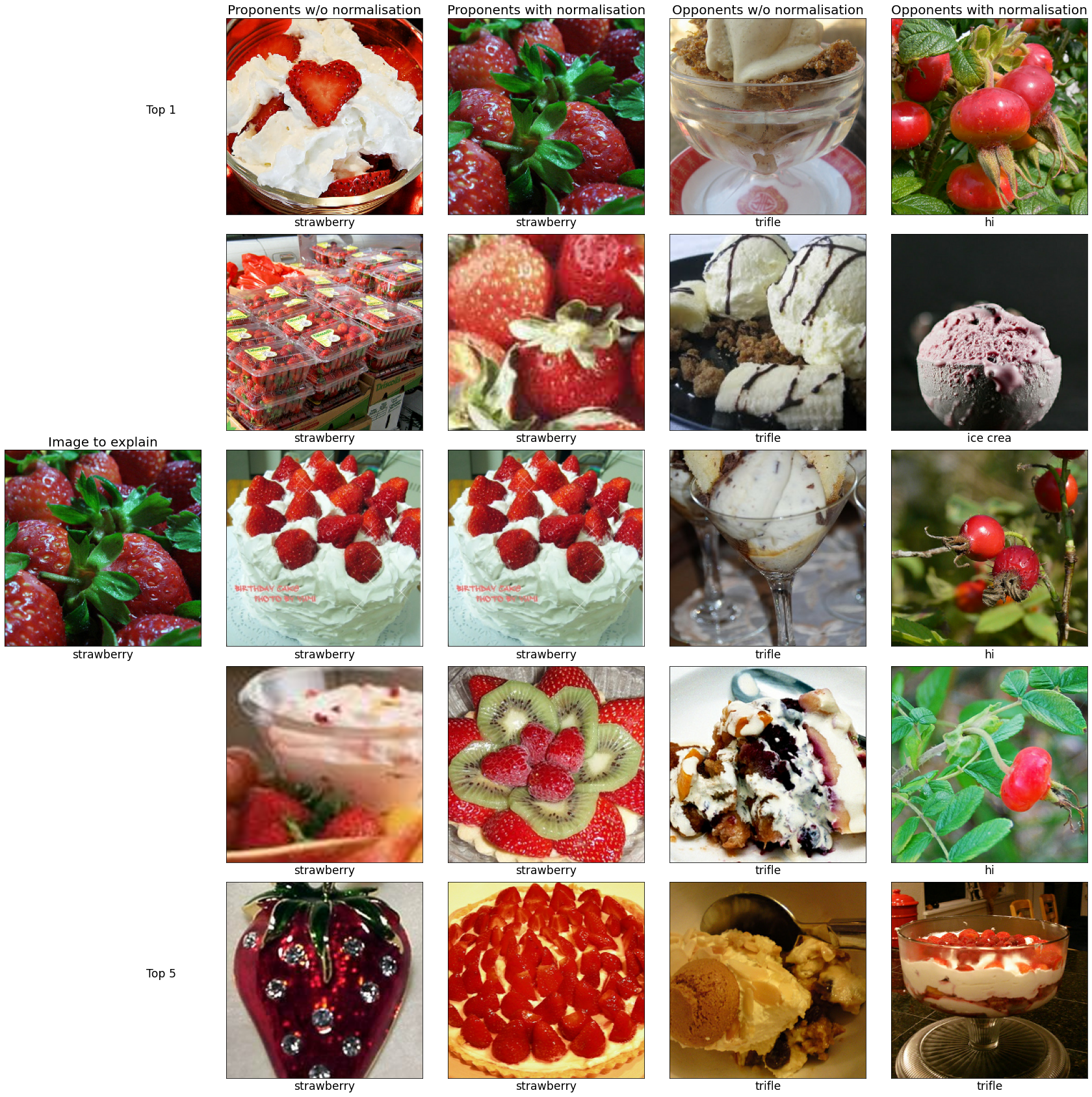}
\caption{Another exemplar of ImageNet picture accompanied by it top 5 proponents and opponents with and without normalisation. The figure demonstrates the effect of normalization on the influence scores. In the first column we can see a lot of complicated images, even though they all contain strawberies, these examples are outliers since they have external objects present or the strawberries are of unusual shape. These examples are useful for model learning, therefore the gradients is high, after normalization, however, the proponent images are less "unusual". We can also see in the second column that the image we are explaining is the highest proponent of itself, which meets our expectations. }
\label{fig:res-net-exemplars_prop_op_2}
\end{figure*}

In Figures~\ref{fig:res-net-exemplars_prop_op_1} and~\ref{fig:res-net-exemplars_prop_op_2} we demonstrate the top-5 proponents and opponents of randomly selected images from the ImageNet train set. For a similar experiment,
the protocol in \citet{tracin} did not use the gradients with respect to the parameters (even when considering the last layer only), but to the
logits. This results in the gradients being simplified to one hot vector predictions minus probabilities in place of gradients, effectively resulting in a pre-filtering looking for proponents in the same
class and opponents in the rest of the top predicted labels. We also used such a pre-filtering,
but used the actual projectors with respect to the model parameters, which makes our experiment more challenging. Additionally, we show retrieved images with normalized gradients used in the definition of influence to illustrate IF's sensitivity to the norm of the gradient (e.g.~on high loss images) and that normalization mitigates this.


\subsection{Memorization vs.~Generalization Trade-Offs}\label{app:mem_gen_tradeoff}
In~\S\ref{app:resnet} we found that high self-influence examples appear (to a human) harder to discern or are possibly mislabeled.
This, however, was a qualitative evaluation, so we further investigated the effect of removing data with low/high self-influence.
\citet{feldman_practical} considered the effect of removing examples with a high value of $\mathcal{I}_{\rm mem}$
and found that this lowers test accuracy. With ImageNet being rather clean, it is expected that
removing data would have a negative effect on accuracy too. However, while the examples with high $\mathcal{I}_H(x,x)$ (or $\mathcal{I}_{\rm mem}$) may require to be ``memorized'', the examples with low self-influence might help with \emph{generalization}, i.e.~give the model a solid classification basis for typical images.

\begin{figure}[h]
    \centering
    \includegraphics[width=\columnwidth]{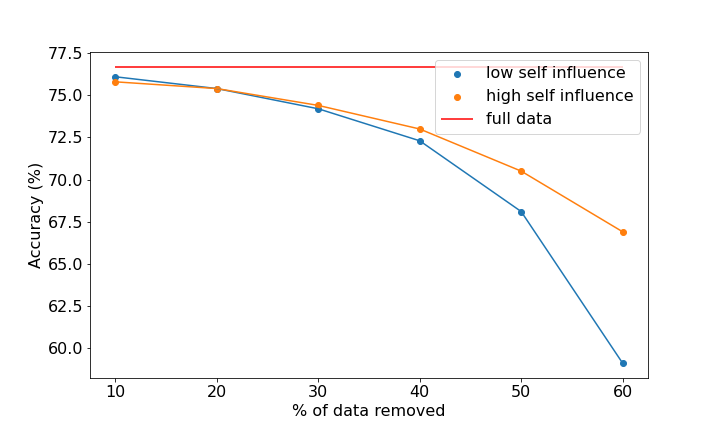}
    \caption{Test accuracy of ResNet50 when progressively removing the most or the least self-influential data.}
    \label{fig:mem_vs_gen_resnet}
\end{figure}

To test the trade-off between memorization and generalization we remove the same percentage of data
either starting from the examples with high or low self-influence and retrain ResNet50. Results are
reported in Figure~\ref{fig:mem_vs_gen_resnet}, where the horizontal line gives the accuracy on training on the
full data. We find that removing examples with low self-influence hurts performance more than
removing those with high self-influence suggesting that in this setting \emph{generalization} has more
incremental value over \emph{memorization}. 

As our estimator for self-influence is less expensive than $\mathcal{I}_{\rm mem}$ we hope that this
can spur further research 1) on investigating the generalization-memorization trade-off in other applications, and 2) on experimenting with training schedules to decide in which order to expose easier or harder data to a learning algorithm~\cite{kreutzer21bandits}.

\end{document}